\documentclass[10pt, a4paper]{article}
\usepackage{lrec}
\usepackage{multibib}
\newcites{languageresource}{Language Resources}
\usepackage{graphicx}
\usepackage{tabularx}
\usepackage{booktabs}
\usepackage{multirow}
\usepackage{fontawesome}
\usepackage{amsmath,calc}
\usepackage{soul}
\usepackage{xcolor}
\usepackage{xspace}
\usepackage{tablefootnote}
\usepackage{array}
\usepackage{ragged2e}
\usepackage{caption}
\usepackage{subcaption}

\usepackage{epstopdf}
\usepackage[utf8]{inputenc}

\usepackage{hyperref}
\usepackage{xstring}

\title{Controllable Sentence Simplification}

\newcommand{\email}[1]{#1}

\author{Louis Martin \\
  Facebook AI Research \& Inria \\
  \email{louismartin@fb.com} \\\And
  Beno{\^\i}t Sagot\\
  Inria\\
  \email{benoit.sagot@inria.fr}\\\And
  \'Eric de la Clergerie\\
  Inria  \\
  \email{eric.de\_la\_clergerie@inria.fr}\\\And
  Antoine Bordes \\
  Facebook AI Research  \\
  \email{abordes@fb.com} \\
  }

\name{Louis Martin$^{1,2,3}$\quad \'Eric Villemonte de la Clergerie$^2$\quad Beno\^it Sagot$^2$\quad Antoine Bordes$^{1}$}

\address{Facebook AI Research$^1$, Inria$^2$, Sorbonne Université$^3$.\\
         6 Rue Ménars, 75002 Paris$^1$; 2 rue Simone Iff, 75012 Paris$^2$.\\
         louismartin@fb.com, \{benoit.sagot, eric.de\_la\_clergerie\}@inria.fr, antoine.bordes@fb.com\\}

\abstract{
    Text simplification aims at making a text easier to read and understand by simplifying grammar and structure while keeping the underlying information identical. It is often considered an all-purpose generic task where the same simplification is suitable for all; however multiple audiences can benefit from simplified text in different ways. We adapt a discrete parametrization mechanism that provides explicit control on simplification systems based on Sequence-to-Sequence models. As a result, users can condition the simplifications returned by a model on attributes such as length, amount of  paraphrasing, lexical complexity and syntactic complexity. We also show that carefully chosen values of these attributes allow out-of-the-box Sequence-to-Sequence models to outperform their standard counterparts on simplification benchmarks. Our model, which we call \textsc{ACCESS} (as shorthand for AudienCe-CEntric Sentence Simplification), establishes the state of the art at 41.87 SARI  on the \mbox{WikiLarge} test set, a +1.42 improvement over the best previously reported score.\\ \newline 
}

\begin{document}

\maketitleabstract

\section{Introduction}

In Natural Language Processing, the Text Simplification task aims at making a text easier to read and understand.
Text simplification can be beneficial for people with cognitive disabilities such as aphasia \cite{carroll1998practical}, dyslexia \cite{rello2013simplify} and autism \cite{evans2014evaluation} but also for second language learners \cite{xia2016text} and people with low literacy \cite{watanabe2009facilita}.
The type of simplification needed for each of these audiences is different.
Some aphasic patients struggle to read sentences with a high cognitive load such as long sentences with intricate syntactic structures, whereas second language learners might not understand texts with rare or specific vocabulary.
Yet, research in text simplification has been mostly focused on developing models that generate a single generic simplification for a given source text with no possibility to adapt outputs for the needs of various target populations.

In this paper, we propose a controllable simplification model that provides explicit ways for users to manipulate and update simplified outputs as they see fit.
This work only considers the task of {\it Sentence Simplification} (SS) where the input of the model is a single source sentence and the output can be composed of one sentence or split into multiple.
Our work builds upon previous work on controllable text generation \cite{kikuchi2016controlling,fan2017controllable,scarton2018learning,nishihara2019controllable} where a Sequence-to-Sequence (Seq2Seq) model is modified to control attributes of the output text.
We tailor this mechanism to the task of SS by considering relevant attributes of the output sentence such as the output length, the amount of paraphrasing, lexical complexity, and syntactic complexity.
To this end, we condition the model at train time, by feeding control tokens representing these attributes along with the source sentence as additional inputs.

Our contributions are the following:
(1) We adapt a parametrization mechanism to the specific task of Sentence Simplification by conditioning on relevant attributes;
(2) We show through a detailed analysis that our model can indeed control the considered attributes, making the simplifications potentially able to fit the needs of various end audiences; (3) With careful calibration, our controllable parametrization improves the performance of out-of-the-box Seq2Seq models leading to a new state-of-the-art score of 41.87 SARI \cite{xu2016optimizing} on the WikiLarge benchmark \citelanguageresource{zhang2017sentenceresource}, a +1.42 gain over previous scores, without requiring any external resource or modified training objective.

\section{Related Work}

\subsection{Sentence Simplification}
Text simplification has gained increasing interest through the years and has benefited from advances in Natural Language Processing and notably Machine Translation.

In recent years, SS was largely treated as a monolingual variant of machine translation (MT), where simplification operations are learned from complex-simple sentence pairs automatically extracted from English Wikipedia and Simple English Wikipedia \cite{zhu2010monolingual,wubben2012sentence}.

Phrase-based and Syntax-based MT was successfully used for SS \cite{zhu2010monolingual} and further tailored to the task using deletion models \cite{coster2011learning} and candidate reranking \cite{wubben2012sentence}.
The candidate reranking method by \newcite{wubben2012sentence} favors simplifications that are most dissimilar to the source using Levenshtein distance. The authors argue that dissimilarity is a key factor of simplification.

Lately, SS has mostly been tackled using Seq2Seq MT models \cite{sutskever2014sequence}.
Seq2Seq models were either used as-is \cite{nisioi2017exploring} or combined with reinforcement learning thanks to a specific simplification reward \cite{zhang2017sentence}, augmented with an external simplification database as a dynamic memory \cite{zhao2018integrating} or trained with multi-tasking on entailment and paraphrase generation \cite{guo2018dynamic}.

This work builds upon Seq2Seq as well. We prepend additional inputs to the source sentences at train time, in the form of plain text control tokens. Our approach does not require any external data or modified training objective.

\subsection{Controllable Text Generation}
Conditional training with Seq2Seq models was applied to multiple natural language processing tasks such as summarization \cite{kikuchi2016controlling,fan2017controllable}, dialog \cite{see2019makes}, sentence compression \cite{fevry2018unsupervised,mallinson2018sentence} or poetry generation \cite{ghazvininejad2017hafez}. 

Most approaches for controllable text generation are either decoding-based or learning-based.

\paragraph{Decoding-based methods}
Decoding-based methods use a standard Seq2Seq training setup but modify the system during decoding to control a given attribute. For instance, the length of summaries was controlled by preventing the decoder from generating the End-Of-Sentence token before reaching the desired length or by only selecting hypotheses of a given length during the beam search \cite{kikuchi2016controlling}.
Weighted decoding (i.e. assigning weights to specific words during decoding) was also used with dialog models \cite{see2019makes} or poetry generation models \cite{ghazvininejad2017hafez} to control the number of repetitions, alliterations, sentiment or style.

\paragraph{Learning-based methods}
On the other hand, learning-based methods condition the Seq2Seq model on the considered attribute at train time, and can then be used to control the output at inference time.
\newcite{kikuchi2016controlling} explored learning-based methods to control the length of summaries, e.g. by feeding a target length vector to the neural network.
They concluded that learning-based methods worked better than decoding-based methods and allowed finer control on the length without degrading performances.
Length control was likewise used in sentence compression by feeding the network a length countdown scalar  \cite{fevry2018unsupervised} or a length vector \cite{mallinson2018sentence}.
\cite{ficler2017controlling} concatenate a context vector to the hidden state of each time step of their recurrent neural network decoder.
This context vector represents the controlled stylistic attributes of the text, where an embedding is learnt for each attribute value.
\cite{hu2017toward} achieved controlled text generation by disentangling the latent space representations of a variational auto-encoder between the text representation and its controlled attributes such as sentiment and tense. They impose the latent space structure during training by using additional discriminators.

Our work uses a simpler approach: we condition the generation process by concatenating plain text control tokens to the source text. This method only modifies the source data and not the training procedure.
Such mechanism was used to control politeness in MT \cite{sennrich2016controlling}, to control summaries in terms of length, of news source style, or to make the summary more focused on a given named entity \cite{fan2017controllable}.
\newcite{scarton2018learning} and \newcite{nishihara2019controllable} similarly showed that adding control tokens at the beginning of sentences can improve the performance of Seq2Seq models for SS. Plain text control tokens were used to encode attributes such as the target school grade-level (i.e. understanding level) and the type of simplification operation applied between the source and the ground truth simplification (identical, elaboration, one-to-many, many-to-one).
Our work goes further by using a more diverse set of control tokens that represent specific grammatical attributes of the text simplification process.
Moreover, we investigate the influence of those control tokens on the generated simplification in a detailed analysis.

\section{Adding Control Tokens to Seq2Seq} \label{method_section}
In this section we present \textsc{ACCESS}, our approach for \textbf{A}udien\textbf{C}e-\textbf{CE}ntric \textbf{S}entence \textbf{S}implification.
We want to control the process of Sentence Simplification using explicit control tokens.
We first identify attributes that cover important aspects of the simplification process and then find explicit control tokens to represent each of those attributes.
Parametrization is then achieved by conditioning a Seq2Seq model on those control tokens.

\subsection{Controlled Attributes}
Based on previous findings, we identify four attributes related to the process of text simplification:  amount of compression, amount of paraphrasing, lexical complexity and syntactic complexity,.

\begin{itemize}
\item
{\bf Amount of compression:}
The amount of compression is directly dependent on the length of sentences which is itself very correlated to simplicity \cite{martin2019reference}, and is one of the two variables used in FKGL \cite{kincaid1975derivation}.
It also accounts for the amount of content that is preserved between the source and target text, and can therefore control the simplicity-adequacy trade-off that is witnessed in text simplification \cite{schwarzer2018human}.

\item
{\bf Paraphrasing:}
Paraphrasing is an important aspect for good text simplification systems \cite{wubben2012sentence}, especially because it allows the user from choosing if he prefers very safe simplifications (i.e. close to the source) or to try and simplify the input more at the cost of more mistakes when using imperfect systems.
The amount of paraphrasing was also shown to correlate with human jugdment of meaning preservation and simplicity sometimes even more than traditional metrics such as BLEU \cite{papineni2002bleu} and SARI \cite{xu2016optimizing}.

\item
{\bf Lexical and Syntactic complexity:}
\cite{shardlow2014survey} identified lexical simplification and syntactic simplification as core components of SS systems, which often decomposes there approach into these two sub-components. Audiences also have different simplification needs along these two attributes. In order to understand a text correctly, second language learner will require a text with less complicated words. On the other hand, some specific types of aphasia will make people struggle more with complex syntactic structures, intricated clauses, and long sentence, thus requiring syntactic simplification.
\end{itemize}

Other more specific attributes could be considered such as the tense or the use passive-active voice. We only consider the previous attributes for simplicity and leave the rest for future work.
We don't consider ``readability'' measured with FKGL because it is just a linear combination of other attributes, namely sentence length and word complexity.

\subsection{Explicit Control Tokens}
For each of the four aforementioned attributes, we choose an explicit ``proxy'' control token that can be computed using the source and simplified sentence and used as a plain text token.
We describe these for explicit control tokens in this subsection.

\begin{itemize}
    \item
{\bf NbChars:} character length ratio between source sentence and target sentence (compression level). This control token accounts for sentence compression, and content deletion.
Previous work showed that simplicity is best correlated with length-based metrics, and especially in terms of number of characters \cite{martin2019reference}. The number of characters indeed accounts for the lengths of words which is itself correlated to lexical complexity.
    \item 
{\bf LevSim:} normalized character-level Levenshtein similarity \cite{levenshtein1966binary} between source and target. LevSim quantifies the amount of modification operated on the source sentence (through paraphrasing, adding and deleting content).
    \item 
{\bf WordRank:} as a proxy to lexical complexity, we compute a sentence-level measure, that we call \mbox{\textit{WordRank}}, by taking the third-quartile of log-ranks (inverse frequency order) of all words in a sentence.  We subsequently divide the \mbox{\textit{WordRank}} of the target by that of the source to get a ratio.
Word frequencies have shown to be the best indicators of word complexity in the Semeval 2016 task 11 \cite{paetzold2016semeval}.
    \item 
{\bf DepTreeDepth:} maximum depth of the dependency tree of the source divided by that of the target (we do not feed any syntactic information other than this ratio to the model). This control token is designed to approximate syntactic complexity. Deeper dependency trees indicate dependencies that span longer and possibly more intricate sentences. DepTreeDepth proved better in early experiments over other candidates for measuring syntactic complexity such as the maximum length of a dependency relation, or the maximum inter-word dependency flux.
\end{itemize}

We parametrize a Seq2Seq model on a given attribute of the target simplification, e.g.~its length, by prepending a control token at the beginning of the source sentence.
The control token value is the ratio\footnote{Early experiments showed that using a ratio instead of an absolute value allowed finer control on the respective attributes.}
of this control token calculated on the target sentence with respect to its value on the source sentence.
For example when trying to control the number of characters of a generated simplification, we compute the compression ratio between the number of characters in the source and the number of characters in the target sentence (see Table~\ref{conditioning_example} for an illustration). Ratios are discretized into bins of fixed width of 0.05 in our experiments and capped to a maximum ratio of 2. Control tokens are then included in the vocabulary (40 unique values per control token).

At inference time, we just set the ratio to a fixed value for all samples\footnote{We did not investigate predicting ratios on a per sentence basis as done by \newcite{scarton2018learning}, and leave this for future work. End-users can nonetheless choose the target ratios as they see fit, for each source sentence.}. For instance, to get simplifications that are 80\% of the source length, we prepend the token $<$NbChars\_0.8$>$ to each source sentence.
This fixed ratio can be user-defined or automatically set.
In our setting, we choose fixed ratios that maximize the SARI on the validation set.

\begin{table}
\small
\centering
\begin{tabular}{p{0.1\columnwidth}p{0.8\columnwidth}}
\toprule
Source& \textit{$<$NbChars\_0.3$>$ $<$LevSim\_0.4$>$  He settled in London~, devoting himself chiefly to practical teaching~.}\\
& \\
Target & \textit{He teaches in London~.}\\
\bottomrule
\end{tabular}
\caption{Example of parametrization on the number of characters. Here the source and target simplifications respectively contain $71$ and $22$ characters which gives a compression ratio of $0.3$. We prepend the $<$NbChars\_0.3$>$ token to the source sentence. Similarly, the Levenshtein similarity between the source and the sentence is 0.37 which gives the $<$LevSim\_0.4$>$ control token after bucketing.}
\label{conditioning_example}
\end{table}

\section{Experiments}

\subsection{Experimental Setting}

\paragraph{Architecture details}
We train a Transformer model \cite{vaswani2017attention} using the FairSeq toolkit \cite{ott2019fairseq}.
Our architecture is the base architecture from \cite{vaswani2017attention}. We used an embedding dimension of 512, fully connected layers of dimension 2048, 8 attention heads, 6 layers in the encoder and 6 layers in the decoder. Dropout is set to 0.2. We use the Adam optimizer \cite{kingma2014adam} with $\beta_1 = 0.9$, $\beta_2 = 0.999$, $\epsilon = 10^{ -8}$ and a learning rate of $lr = 0.00011$. We add label smoothing with a uniform prior distribution of $\epsilon = 0.54$. We use early stopping when SARI does not increase for more than 5 epochs. We tokenize sentences using the NLTK NIST tokenizer and preprocess using SentencePiece \cite{kudo2018sentencepiece} with 10k vocabulary size to handle rare and unknown words. For generation we use beam search with a beam size of 8.
\footnote{Code and pretrained models are released with an open-source license at \mbox{\url{https://github.com/facebookresearch/access}}.}

\paragraph{Training and evaluation datasets}
Our models are trained and evaluated on the \textbf{WikiLarge dataset} \citelanguageresource{zhang2017sentenceresource} which contains 296,402/2,000/359 samples (train/validation/test).
WikiLarge is a set of automatically aligned complex-simple sentence pairs from English Wikipedia (EW) and Simple English Wikipedia (SEW). It is compiled from previous extractions of EW-SEW \cite{zhu2010monolingual,woodsend2011learning,kauchak2013improving}.
Its validation and test sets are taken from Turkcorpus \citelanguageresource{xu2016optimizingresource}, where each complex sentence has 8 human simplifications created by Amazon Mechanical Turk workers.
Human annotators were instructed to only paraphrase the source sentences while keeping as much meaning as possible. Hence, no sentence splitting, minimal structural simplification and little content reduction occurs in this test set \cite{xu2016optimizing}.
We are not able to use the Newsela dataset \citelanguageresource{xu2015problemsresource} because of legal constraints related to its limited public availability. The Newsela dataset can only be accessed by signing a one year Data Sharing Agreement and comes with a restrictive non-commercial license.
Additionally, all publications using the dataset need to be sent in advance to Newsela for approval.
This limited public availability also prevents the research community from agreeing on a public train/validation/test split which hampers reproducibility of results.

\paragraph{Evaluation metrics}
We evaluate our methods with \textbf{FKGL} (Flesch-Kincaid Grade Level)  \cite{kincaid1975derivation} to account for simplicity and \textbf{SARI} \cite{xu2016optimizing} as an overall score.
FKGL is a commonly used metric for measuring readability however it should not be used alone for evaluating systems because it does not account for grammaticality and meaning preservation \cite{wubben2012sentence}. It is computed as a linear combination of the number of words per simple sentence and the number of syllables per word:

\[FKGL = 0.39\frac{nb\ words}{nb\ sentences} + 11.8\frac{nb\ syllables}{nb\ words} - 15.59\]

On the other hand SARI compares the predicted simplification with both the source \textit{and} the target references. It is an average of F1 scores for three $n$-gram operations: additions, keeps and deletions\footnote{Following \newcite{zhang2017sentence}, our SARI implementation includes deletion recall to match previous work.}. For each operation, these scores are then averaged for all $n$-gram orders (from 1 to 4) to get the overall F1 score.

    \[ope \in [add, keep, del]\]
    \[f_{ope}(n) = \frac{2 \times p_{ope}(n) \times r_{ope}(n)}{p_{ope}(n)+r_{ope}(n)}\]
    \[F_{ope} = \frac{1}{k}\sum_{n=[1,..,k]}f_{ope}(n)\]
    \[SARI = \frac{F_{add} + F_{keep} + F_{del}}{3}\]

We compute FKGL and SARI using the EASSE python package for SS \cite{alva2019easse}.
We do not use BLEU because it is not suitable for evaluating SS systems \cite{sulem2018bleu}.
BLEU is also misleading because it favors models that do not modify the source sentence \cite{xu2016optimizing} on TurkCorpus. For instance copying the source sentence in place of simplification gives a BLEU of 99.37 on WikiLarge.

\subsection{Overall Performance}

Table~\ref{comparison_to_sota} compares our best model to state-of-the-art methods:
\begin{description}
  \item [PBMT-R] \cite{wubben2012sentence}\\
  Phrase-Based MT system with candidate reranking. Dissimilar candidates are favored based on their Levenshtein distance to the source. 
  \item [Hybrid] \cite{narayan2014hybrid}\\
  Deep semantics sentence representation fed to a monolingual MT system.
  \item [SBMT+PPDB+SARI] \cite{xu2016optimizing}\\ Syntax-based MT model augmented using the PPDB paraphrase database \cite{pavlick2015ppdb} and fine-tuned towards SARI. 
  \item [DRESS-LS] \cite{zhang2017sentence}\\
  Seq2Seq trained with reinforcement learning, combined with a lexical simplification model. 
  \item [Pointer+Ent+Par] \cite{guo2018dynamic}\\
  Seq2Seq model based on the pointer-copy mechanism and trained via multi-task learning on the Entailment and Paraphrase Generation tasks. 
  \item [NTS+SARI] \cite{nisioi2017exploring}\\
  Standard Seq2Seq model. The second beam search hypothesis is selected during decoding; the hypothesis number is an hyper-parameter fine-tuned with SARI. 
  \item [NSELSTM-S] \cite{vu2018sentence}\\
  Seq2Seq with a memory-augmented Neural Semantic Encoder, tuned with SARI. 
  \item [DMASS+DCSS] \cite{zhao2018integrating}\\
  Seq2Seq integrating the simple PPDB simplification database \cite{pavlick2016simple} as a dynamic memory. The database is also used to modify the loss and re-weight word probabilities to favor simpler words.
\end{description}

\begin{table}
\resizebox{\columnwidth}{!}{
    \begin{tabular}{p{0.65\linewidth}cc}
\toprule
WikiLarge \textbf{(test)}                                                                               & SARI $\uparrow$ & FKGL $\downarrow$ \\
\midrule
PBMT-R        & $38.56$           & $8.33$             \\
Hybrid                                                                                                  & $31.40 $          & $\mathbf{4.56}$     \\
SBMT+PPDB+SARI                                                                                          & $39.96$           & $7.29$              \\
DRESS-LS                                                                                                & $37.27$           & $6.62$              \\
Pointer+Ent+Par                                                                                         & $37.45$           & ---               \\
NTS+SARI                                                                                              & $37.25$           & ---               \\
NSELSTM-S                                                                                               & $36.88$           & ---               \\
DMASS+DCSS                                                                                            & $40.45$           & $8.04$              \\
\midrule
{\bf ACCESS:} NbChars$_{0.95}$ + LevSim$_{0.75}$ + WordRank$_{0.75}$ & \textbf{41.87}  & $7.22$  \\
\bottomrule
    \end{tabular}
}
\caption{Comparison to the literature. We report the results of the model that performed the best on the validation set among all runs and parametrizations. The ratios used for parametrizations are written as subscripts.}
\label{comparison_to_sota}
\end{table}

We select the model with the best SARI on the validation set and report its score on the test set. This model uses three control tokens out of four: NbChars$_{0.95}$, LevSim$_{0.75}$ and WordRank$_{0.75}$ (optimal target ratios in subscript).

\textsc{ACCESS} scores best on SARI (41.87), a significant improvement over previous state of the art (40.45), and third to best FKGL (7.22).
The second and third models in terms of SARI, DMASS+DCSS (40.45) and SBMT+PPDB+SARI (39.96), both use the external resource Simple PPDB \cite{pavlick2016simple} that was extracted from 1000 times more data than what we used for training. Our FKGL is also better (lower) than these methods.
The Hybrid model scores best on FKGL (4.56) i.e.~they generated the simplest (and shortest) sentences, but it was done at the expense of SARI (31.40).

\begin{figure}
    \centering
    \includegraphics[width=\columnwidth]{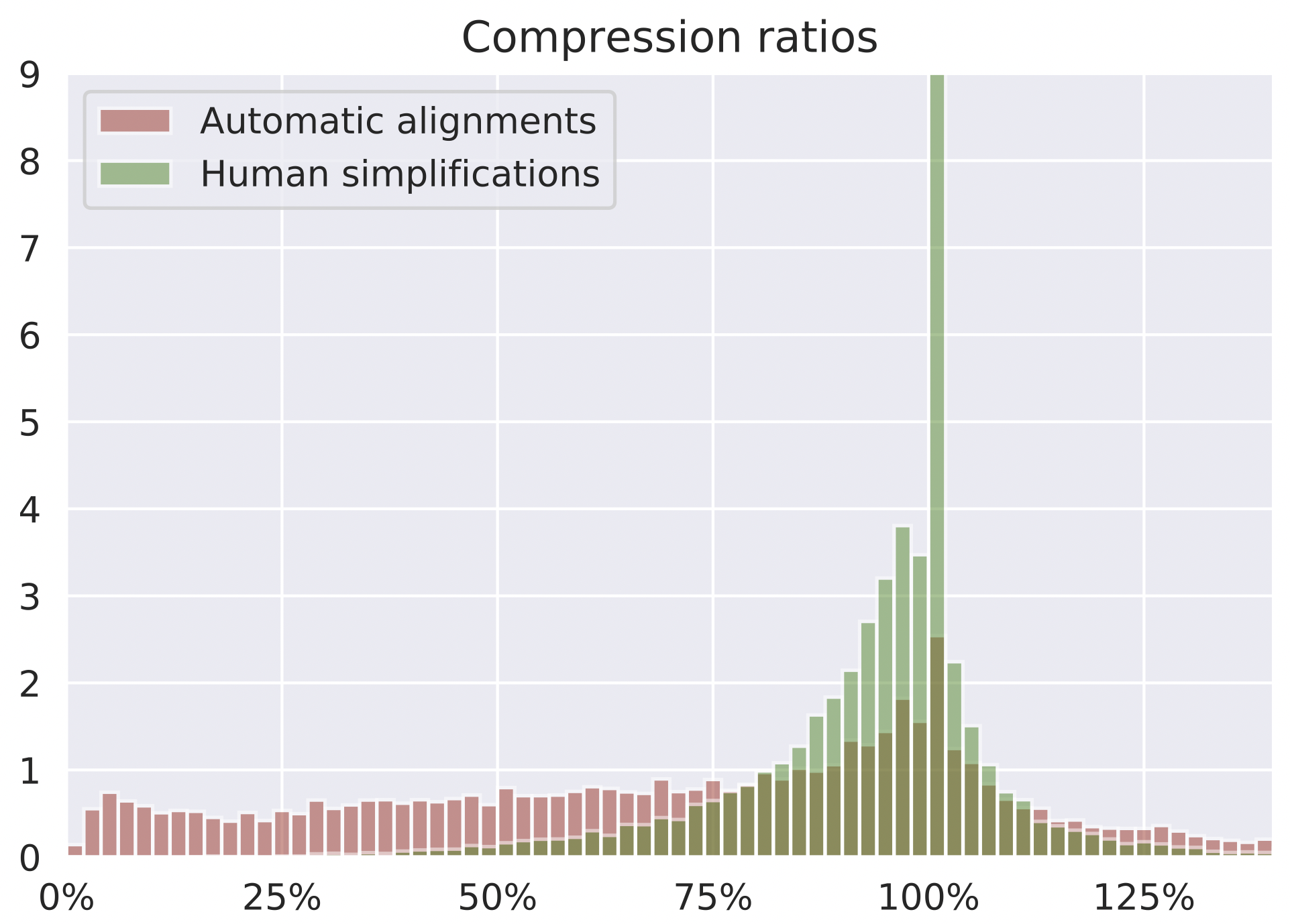}
        \caption{Density distribution of the \textbf{compression ratios} between the source sentence and the target sentence. The automatically aligned pairs from WikiLarge train set are spread (red) while human simplifications from the validation and test set (green) are gathered together with a mean ratio of 0.93 (i.e. nearly no compression). \label{automatic_vs_human}}
\end{figure}

Parametrization encourages the model to rely on explicit aspects of the simplification process, and to associate them with the control tokens. The model can then be adapted more precisely to the type of simplification needed.
In WikiLarge, for instance, the compression ratio distribution is different than that of human simplifications (see Figure~\ref{automatic_vs_human}).
The NbChars control token helps the model decorrelate the compression aspect from other attributes of the simplification process. This control token is then adapted to the amount of compression required in a given evaluation dataset, such as a true, human simplified SS dataset.
Our best model indeed worked best with a NbChars target ratio set to 0.95 which is the closest bucketed value to the compression ratio of human annotators on the WikiLarge validation set (0.93).

\section{Ablation Studies}
\begin{table}
\small
\begin{tabular}{p{0.4\columnwidth}ll}
\toprule
WikiLarge \textbf{(validation)} & SARI $\uparrow$           & FKGL $\downarrow$        \\
\midrule
    
Transformer & $37.06 \pm 0.25$ & $7.66 \pm 0.42$ \\
\midrule
+DepTreeDepth & $37.72^{*} \pm 0.18$ & $7.64 \pm 0.22$ \\
+NbChars & $37.94^{*} \pm 0.09$ & $7.87 \pm 0.15$ \\
+LevSim & $38.29^{*}  \pm 0.66$ & $7.53 \pm 0.21$ \\
+WordRank & $39.35^{*} \pm 0.25$ & $7.61 \pm 0.19$ \\
\midrule
+WordRank+LevSim & $41.1^{*} \pm 0.14$ & $6.86^{*} \pm 0.17$ \\
\midrule
+WordRank+LevSim +NbChars & $\mathbf{41.29^{*}} \pm 0.27$ & $7.25^{*} \pm 0.26$ \\
\midrule
\textit{all} & $41.03^{*} \pm 0.39$ & $\mathbf{6.72^{*}} \pm 0.39$ \\

\bottomrule
\end{tabular}
\caption{Ablation study on the control tokens using greedy forward selection. We report SARI and FKGL on WikiLarge \textbf{validation} set. Each score is a mean over 10 runs with a 95\% confidence interval. Scores with $^{*}$ are statistically significantly better than the Transformer baseline (p-value $< 0.01$ for a Student's T-test).} \label{ablation}
\end{table}

In this section we investigate the contribution of each control token to the final SARI score of \textsc{ACCESS}.
Table~\ref{ablation} reports scores of models trained with different combinations of control tokens on the WikiLarge validation set (2000 source sentences, with 8 human simplifications each). We combined control tokens using greedy forward selection; at each step, we add the control token leading to the best performance when combined with previously added control tokens.

With only one control token, WordRank proves to be best (+2.28 SARI over models without parametrization).
As the WikiLarge validation set mostly contains small paraphrases, it seems natural that the control token linked to lexical simplification increases the performance the most.

LevSim (+1.23) is the second best control token. This confirms the intuition that hypotheses that are more dissimilar to the source are better simplifications, as claimed in \cite{wubben2012sentence,nisioi2017exploring}.

There is little content reduction in the WikiLarge validation set (see Figure~\ref{automatic_vs_human}), thus control tokens that are closely related to sentence length will be less effective. This is the case for the NbChars and DepTreeDepth control tokens (shorter sentences, will have lower tree depths): they bring more modest improvements, +0.88 and +0.66.

The performance boost is nearly additive at first when adding more control tokens (WordRank+LevSim: +4.04) but saturates quickly with 3+ control tokens. In fact, no combination of 3 or more control tokens gets a statistically significant improvement over the WordRank+LevSim setup (p-value $< 0.01$ for a Student's T-test).
This indicates that control tokens are not all useful to improve the scores on this benchmark, and that they might be not independent from one another.
The addition of the DepTreeDepth as a final control token even decreases the SARI score slightly, most probably because the considered validation set does not include sentence splitting and structural modifications.

\section{Analysis of the Influence of Control Tokens}

\begin{figure*}
    \begin{subfigure}[!htbb]{\textwidth}
        \centering
        \includegraphics[width=0.9\textwidth]{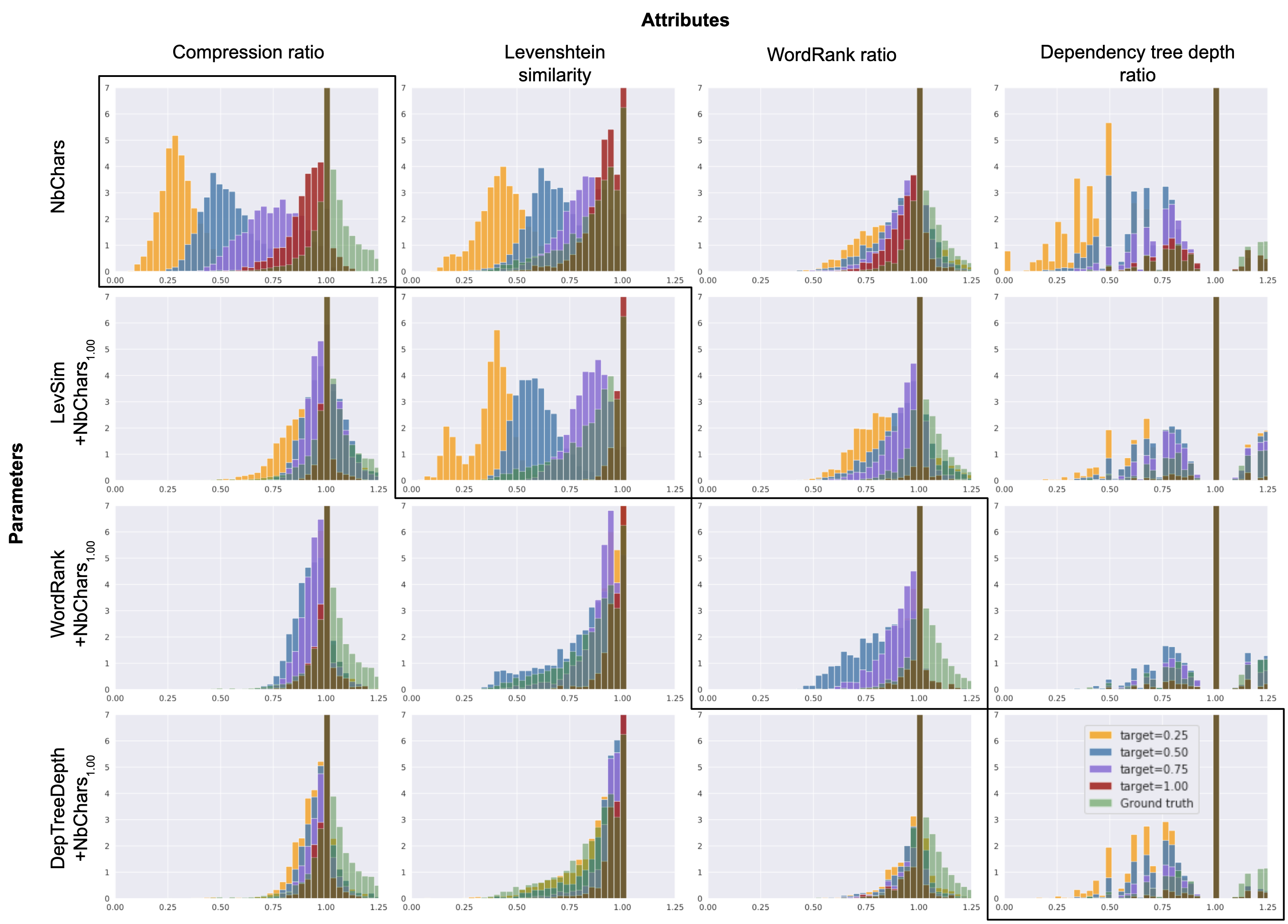}
        \caption{With the NbChars$_{1.00}$ constraint.\label{influence_of_control_tokens_with}}
    \end{subfigure}
    \begin{subfigure}[!htbp]{\textwidth}
        \centering
        \includegraphics[width=0.9\textwidth]{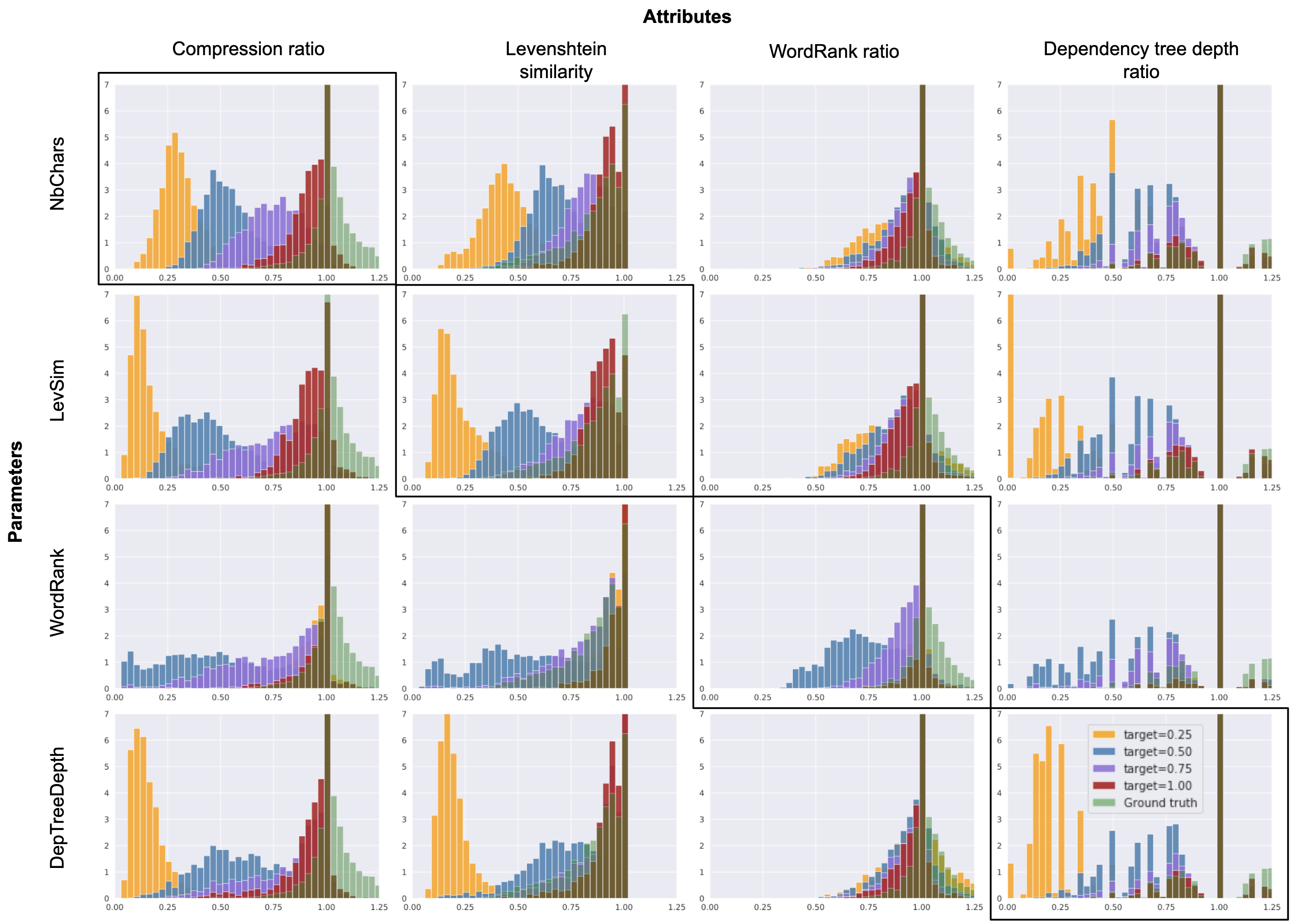}
        \caption{Without the NbChars$_{1.00}$ constraint.\label{influence_of_control_tokens_without}}
    \end{subfigure}
    \caption{Influence of each control token on the corresponding attributes of the output simplifications. \textbf{Rows represent control tokens} (each model is trained either only with one control token or with one control token and the NbChars$_{1.00}$ constraint), \textbf{columns represent output attributes} of the predictions and \textbf{colors represent the fixed target ratio} of the control token (yellow=0.25, blue=0.50, violet=0.75, red=1.00, green=Ground truth). We plot the results on the 2000 validation sentences. Figure~\ref{influence_of_control_tokens_with} uses the  NbChars$_{1.00}$ constraint, whereas Figure~\ref{influence_of_control_tokens_without} doesn't.\label{influence_of_control_tokens}}
\end{figure*}

\begin{table*}
    \centering
\resizebox{\textwidth}{!}{
    \begin{tabular}{ll}
    \toprule
     Target control tokens & Sentence \\
\midrule
\textbf{Source} & Some trails are designated as nature trails , and are used by people learning about the natural world . \\
\midrule
\textbf{NbChars$_{\bf 1.00}$} & Some trails are \textbf{called} nature trails , and are used by people about the natural world . \\
\textbf{NbChars$_{\bf 0.75}$} & Some trails are \textbf{called} nature trails , and are used by people about the natural world . \\
\textbf{NbChars$_{\bf 0.50}$} & Some trails are used by people about the natural world . \\
\textbf{NbChars$_{\bf 0.25}$} & Some trails are used by people . \\
\midrule
\textbf{LevSim$_{\bf 1.00}$}\footnotesize{+NbChars$_{1.00}$} & Some trails are designated as nature trails , and are used by people learning about the natural world . \\
\textbf{LevSim$_{\bf 0.75}$}\footnotesize{+NbChars$_{1.00}$} & Some trails are \textbf{made} \textbf{for} nature trails \textbf{.} \textbf{They} are used by people \textbf{who} \textbf{learn} about the natural world . \\
\textbf{LevSim$_{\bf 0.50}$}\footnotesize{+NbChars$_{1.00}$} & \textbf{The} trails \textbf{that} are used by people learning about the natural world \textbf{,} \textbf{because} \textbf{the} \textbf{trails} \textbf{are} \textbf{good} \textbf{trails} . \\
\textbf{LevSim$_{\bf 0.25}$}\footnotesize{+NbChars$_{1.00}$} & \textbf{Mechanical} trails \textbf{(} \textbf{also} \textbf{known} \textbf{as} \textbf{''} \textbf{trail} \textbf{trail} \textbf{''} \textbf{or} \textbf{''} \textbf{trails} \textbf{''} \textbf{)} are trails \textbf{that} are used \textbf{for} \textbf{trails} . \\
\midrule
\textbf{WordRank$_{\bf 1.00}$}\footnotesize{+NbChars$_{1.00}$} & Some trails are designated as nature trails , and are used by people learning about the natural world . \\
\textbf{WordRank$_{\bf 0.75}$}\footnotesize{+NbChars$_{1.00}$} & Some trails are \textbf{called} nature trails , and are used by people learning about the natural world . \\
\textbf{WordRank$_{\bf 0.50}$}\footnotesize{+NbChars$_{1.00}$} & Some trails are \textbf{known} as nature trails , and are used by people \textbf{as} \textbf{well} \textbf{as} \textbf{by} \textbf{people} \textbf{who} \textbf{are} \textbf{in} the world . \\
\textbf{WordRank$_{\bf 0.25}$}\footnotesize{+NbChars$_{1.00}$} & Some trails are \textbf{also} \textbf{called} nature trails , and are used by people learning about the natural world . \\
\midrule
\textbf{DepTreeDepth$_{\bf 1.00}$}\footnotesize{+NbChars$_{1.00}$} & Some trails are designated as nature trails , and are used by people learning about the natural world . \\
\textbf{DepTreeDepth$_{\bf 0.75}$}\footnotesize{+NbChars$_{1.00}$} & Some trails are designated as nature trails \textbf{.} \textbf{They} are used by people learning about the natural world . \\
\textbf{DepTreeDepth$_{\bf 0.50}$}\footnotesize{+NbChars$_{1.00}$} & Some trails are designated as nature trails \textbf{.} \textbf{They} are used by people learning about the natural world . \\
\textbf{DepTreeDepth$_{\bf 0.25}$}\footnotesize{+NbChars$_{1.00}$} & Some trails are designated as nature trails \textbf{.} \textbf{They} are used by people \textbf{to} \textbf{learn} about the natural world . \\
\midrule
& \\ \midrule

\textbf{Source} & Iron Maiden , released on April 14 , 1980 , is the debut album by heavy metal band Iron Maiden . \\
\midrule
\textbf{NbChars$_{1.00}$} & Iron Maiden , released on April 14 , 1980 , is the debut album by heavy metal band Iron Maiden . \\
\textbf{NbChars$_{0.75}$} & Iron Maiden is the debut album by heavy metal band Iron Maiden . \\
\textbf{NbChars$_{0.50}$} & Iron Maiden , released on April 14 , 1980 . \\
\textbf{NbChars$_{0.25}$} & Iron Maiden \textbf{was} released on April 14 , 1980 . \\
\midrule
\textbf{LevSim$_{1.00}$}\footnotesize{+NbChars$_{1.00}$} & Iron Maiden , released on April 14 , 1980 , is the debut album by heavy metal band Iron Maiden . \\
\textbf{LevSim$_{0.75}$}\footnotesize{+NbChars$_{1.00}$} & Iron Maiden is the debut album by heavy metal band Iron Maiden . \textbf{It} \textbf{was} \textbf{released} \textbf{on} \textbf{April} \textbf{14} \textbf{,} \textbf{1980} \textbf{.} \\
\textbf{LevSim$_{0.50}$}\footnotesize{+NbChars$_{1.00}$} & Iron Maiden is the debut album by heavy metal band Iron Maiden . \textbf{It} \textbf{was} \textbf{released} \textbf{on} \textbf{April} \textbf{14} \textbf{,} \textbf{1980} \textbf{.} \\
\textbf{LevSim$_{0.25}$}\footnotesize{+NbChars$_{1.00}$} & Iron Maiden is \textbf{an} album by \textbf{American} heavy metal band Iron Maiden . \textbf{It} \textbf{was} \textbf{released} \textbf{on} \textbf{April} \textbf{14} \textbf{,} \textbf{1980} \textbf{.} \\
\midrule
\textbf{WordRank$_{1.00}$}\footnotesize{+NbChars$_{1.00}$} & Iron Maiden is the \textbf{first} album \textbf{released} by heavy metal band Iron Maiden . \textbf{It} \textbf{was} \textbf{released} \textbf{in} \textbf{1980} \textbf{.} \\
\textbf{WordRank$_{0.75}$}\footnotesize{+NbChars$_{1.00}$} & Iron Maiden \textbf{is} \textbf{a} \textbf{first} \textbf{album} \textbf{by} \textbf{the} \textbf{band} \textbf{Iron} \textbf{Maiden} \textbf{.} \textbf{It} \textbf{was} released on April 14 , 1980 . \\
\textbf{WordRank$_{0.50}$}\footnotesize{+NbChars$_{1.00}$} & Iron Maiden is \textbf{a} \textbf{city} \textbf{of} the \textbf{state} \textbf{of} \textbf{Arkansas} \textbf{in} \textbf{the} \textbf{United} \textbf{States} \textbf{of} \textbf{America} . \\
\textbf{WordRank$_{0.25}$}\footnotesize{+NbChars$_{1.00}$} & Iron Maiden \textbf{is} \textbf{a} \textbf{first} \textbf{album} released \textbf{by} \textbf{the} \textbf{band} \textbf{Iron} \textbf{Maiden} \textbf{.} \textbf{It} \textbf{was} \textbf{released} on April 14 , 1980 . \\
\midrule
\textbf{DepTreeDepth$_{1.00}$}\footnotesize{+NbChars$_{1.00}$} & Iron Maiden , released on April 14 , 1980 , is the \textbf{first} album by heavy metal band Iron Maiden . \\
\textbf{DepTreeDepth$_{0.75}$}\footnotesize{+NbChars$_{1.00}$} & Iron Maiden is \textbf{a} \textbf{first} album by \textbf{British} heavy metal band Iron Maiden . \textbf{It} \textbf{was} \textbf{released} \textbf{on} \textbf{April} \textbf{14} \textbf{,} \textbf{1980} \textbf{.} \\
\textbf{DepTreeDepth$_{0.50}$}\footnotesize{+NbChars$_{1.00}$} & Iron Maiden is \textbf{an} album by \textbf{British} heavy metal band Iron Maiden . \textbf{It} \textbf{was} \textbf{released} \textbf{on} \textbf{April} \textbf{14} \textbf{,} \textbf{1980} \textbf{.} \\
\textbf{DepTreeDepth$_{0.25}$}\footnotesize{+NbChars$_{1.00}$} & Iron Maiden \textbf{was} released on April 14 , 1980 \textbf{.} \textbf{It} \textbf{was} \textbf{released} \textbf{in} Iron Maiden \textbf{on} \textbf{April} \textbf{14} \textbf{,} \textbf{1980} . \\
\midrule
& \\ \midrule
\textbf{Source} & Nocturnes is an orchestral composition in three movements by the French composer Claude Debussy . \\
\midrule
\textbf{NbChars$_{1.00}$} & Nocturnes is an orchestral composition in three movements by the French composer Claude Debussy . \\
\textbf{NbChars$_{0.75}$} & Nocturnes is an orchestral composition in three movements by the French composer Claude \textbf{Debus} . \\
\textbf{NbChars$_{0.50}$} & Nocturnes is an orchestral composition in three movements . \\
\textbf{NbChars$_{0.25}$} & Nocturnes is an orchestral composition . \\
\midrule
\textbf{LevSim$_{1.00}$}\footnotesize{+NbChars$_{1.00}$} & Nocturnes is an orchestral composition in three movements by the French composer Claude Debussy . \\
\textbf{LevSim$_{0.75}$}\footnotesize{+NbChars$_{1.00}$} & Nocturnes is \textbf{a} \textbf{piece} \textbf{of} \textbf{music} \textbf{for} \textbf{orchestra} by the French composer Claude Debussy . \\
\textbf{LevSim$_{0.50}$}\footnotesize{+NbChars$_{1.00}$} & Nocturnes is \textbf{a} \textbf{piece} \textbf{of} \textbf{music} \textbf{for} \textbf{orchestra} \textbf{that} \textbf{was} \textbf{composed} by \textbf{a} French composer \textbf{called} Claude Debussy . \\
\textbf{LevSim$_{0.25}$}\footnotesize{+NbChars$_{1.00}$} & \textbf{Claude} \textbf{Debussy} \textbf{was} \textbf{a} French composer \textbf{who} \textbf{wrote} \textbf{music} \textbf{for} \textbf{the} \textbf{orchestra} \textbf{when} \textbf{he} \textbf{was} \textbf{17} \textbf{years} \textbf{old} . \\
\midrule
\textbf{WordRank$_{1.00}$}\footnotesize{+NbChars$_{1.00}$} & Nocturnes is an orchestral composition in three movements by the French composer Claude Debussy . \\
\textbf{WordRank$_{0.75}$}\footnotesize{+NbChars$_{1.00}$} & Nocturnes is \textbf{a} \textbf{piece} \textbf{of} \textbf{music} \textbf{for} \textbf{orchestra} by the French composer Claude Debussy . \\
\textbf{WordRank$_{0.50}$}\footnotesize{+NbChars$_{1.00}$} & Nocturnes is \textbf{a} \textbf{piece} \textbf{of} \textbf{music} by the French composer Claude Debussy . \\
\textbf{WordRank$_{0.25}$}\footnotesize{+NbChars$_{1.00}$} & Nocturnes is \textbf{a} \textbf{piece} \textbf{of} \textbf{music} \textbf{for} \textbf{orchestra} by the French composer Claude Debussy . \\
\midrule
\textbf{DepTreeDepth$_{1.00}$}\footnotesize{+NbChars$_{1.00}$} & Nocturnes is an orchestral composition in three movements by the French composer Claude Debussy . \\
\textbf{DepTreeDepth$_{0.75}$}\footnotesize{+NbChars$_{1.00}$} & Nocturnes is an orchestral composition in three movements by the French composer Claude Debussy . \\
\textbf{DepTreeDepth$_{0.50}$}\footnotesize{+NbChars$_{1.00}$} & Nocturnes is an orchestral composition by the French composer Claude Debussy \textbf{in} \textbf{three} \textbf{movements} . \\
\textbf{DepTreeDepth$_{0.25}$}\footnotesize{+NbChars$_{1.00}$} & Nocturnes is \textbf{a} \textbf{French} \textbf{orchestra} \textbf{.} \textbf{It} \textbf{was} \textbf{started} \textbf{by} \textbf{Claude} \textbf{Debussy} in three movements . \\
\midrule
& \\
\midrule
\textbf{Source} & It is an F-type asteroid , which means that it is very dark in colouring ( darker than soot ) with a carbonaceous composition . \\
\midrule
\textbf{NbChars$_{1.00}$} & It is an F-type asteroid , which means that it is very dark in colouring ( darker than soot ) with a carbonaceous composition . \\
\textbf{NbChars$_{0.75}$} & It is an F-type asteroid , which means that it is very dark in colouring ( darker than soot ) . \\
\textbf{NbChars$_{0.50}$} & \textbf{This} means that it is very dark in colouring ( darker than soot ) . \\
\textbf{NbChars$_{0.25}$} & It is an F-type asteroid . \\
\midrule
\textbf{LevSim$_{1.00}$}\footnotesize{+NbChars$_{1.00}$} & It is an F-type asteroid , which means that it is very dark in colouring ( darker than soot ) with a carbonaceous composition . \\
\textbf{LevSim$_{0.75}$}\footnotesize{+NbChars$_{1.00}$} & It is an F-type asteroid , which means that it is very dark in colouring ( darker than soot ) \textbf{made} \textbf{up} \textbf{of} \textbf{carbonate} \textbf{metal} . \\
\textbf{LevSim$_{0.50}$}\footnotesize{+NbChars$_{1.00}$} & F-type \textbf{asteroids} \textbf{can} \textbf{be} \textbf{made} \textbf{up} \textbf{of} \textbf{darker} \textbf{than} \textbf{soot} \textbf{(} \textbf{darker} \textbf{than} \textbf{soot} \textbf{)} , \textbf{or} \textbf{darker} ( darker than soot ) \textbf{,} \textbf{or} \textbf{dark} \textbf{(} \textbf{darker} \textbf{)} . \\
\textbf{LevSim$_{0.25}$}\footnotesize{+NbChars$_{1.00}$} & \textbf{IAUC} \textbf{2003} \textbf{September} \textbf{6} ( \textbf{naming} \textbf{the} \textbf{moon} ) \textbf{was} \textbf{discovered} \textbf{by} \textbf{Eros} \textbf{in} \textbf{2005} \textbf{by} \textbf{E.} \textbf{H.} \textbf{E.} \textbf{E.} \textbf{J.} \textbf{E.} \textbf{J.} \textbf{J.} \textbf{J.} \textbf{J.} \textbf{J.} \textbf{J.} \textbf{J.} \textbf{R.} \textbf{J.} [...] \\
\midrule
\textbf{WordRank$_{1.00}$}\footnotesize{+NbChars$_{1.00}$} & It is an F-type asteroid , which means that it is very dark in colouring ( darker than soot ) with a carbonaceous composition . \\
\textbf{WordRank$_{0.75}$}\footnotesize{+NbChars$_{1.00}$} & It is an F-type asteroid , which means that it is very dark in colouring ( darker than soot ) with a \textbf{made} \textbf{of} \textbf{carbonate} . \\
\textbf{WordRank$_{0.50}$}\footnotesize{+NbChars$_{1.00}$} & It is an F-type asteroid , which means that it is very dark in colouring ( darker than soot ) with a \textbf{very} \textbf{dark} \textbf{made} \textbf{up} \textbf{of} . \\
\textbf{WordRank$_{0.25}$}\footnotesize{+NbChars$_{1.00}$} & It is an F-type asteroid , which means that it is very dark in colouring ( darker than soot ) with a carbonaceous composition . \\
\midrule
\textbf{DepTreeDepth$_{1.00}$}\footnotesize{+NbChars$_{1.00}$} & It is an F-type asteroid , which means that it is very dark in colouring ( darker than soot ) with a carbonaceous composition . \\
\textbf{DepTreeDepth$_{0.75}$}\footnotesize{+NbChars$_{1.00}$} & It is an F-type asteroid , which means that it is very dark in colouring ( darker than soot ) with a carbonaceous composition . \\
\textbf{DepTreeDepth$_{0.50}$}\footnotesize{+NbChars$_{1.00}$} & It is an F-type asteroid \textbf{.} \textbf{It} means that it is very dark in colouring ( darker than soot ) with a carbonaceous composition . \\
\textbf{DepTreeDepth$_{0.25}$}\footnotesize{+NbChars$_{1.00}$} & It is an F-type asteroid \textbf{.} \textbf{It} means that it is very dark in colouring ( darker than soot ) with a carbonaceous composition . \\
    \bottomrule

    \end{tabular}}
    \caption{Influence of control tokens on example sentences. Each source sentence is simplified with models trained with each of the four control tokens with varying target ratios; modified words are in bold. The NbChars$_{1.00}$ constraint is added for LevSim, WordRank and DepTreeDepth.
    \label{influence_of_control_tokens_on_examples}}
\end{table*}

Our goal is to give the user control over how the model will simplify sentences on four important attributes of SS: length, paraphrasing, lexical complexity and syntactic complexity.
To this end, we introduced four control tokens: NbChars, LevSim, WordRank and DepTreeDepth.
Even though the control tokens improve the performance in terms of SARI, it is not sure whether they have the desired effect on their associated attribute.
In this section we investigate to what extent each control token controls the generated simplification.
We first used separate models, each trained with a single control token to isolate their respective influence on the output simplifications.
However, we witnessed that with only one control token, the effect of LevSim, WordRank and DepTreeDepth was mainly to reduce the length of the sentence (Figure~\ref{influence_of_control_tokens_without}). Indeed, shortening the sentence will decrease the Levenshtein similarity, decrease the WordRank (when complex words are deleted) and decrease the dependency tree depth (shorter sentences have shallower dependency trees).
Therefore, to clearly study the influence of those control tokens, we also add the NbChars control token during training, and set its ratio to 1.00 at inference time, as a constraint toward not modifying the length.

Figure~\ref{influence_of_control_tokens_with} highlights the cross influence of each of the four control tokens on their four associated attributes. Control tokens are successively set to ratios of 0.25 (yellow), 0.50 (blue), 0.75 (violet) and 1.00 (red); the ground truth is displayed in green.
Plots located on the diagonal show that control tokens control their respective attributes (e.g. NbChars affects the compression ratio), although not with the same effectiveness.

The histogram located at (row 1, col 1) shows the effect of the NbChars control token on the compression ratio of the predicted simplifications.
The resulting distributions are centered on the 0.25, 0.5, 0.75 and 1 target ratios as expected, and with little overlap.
This indicates that the lengths of predictions closely follow what is asked of the model. Table~\ref{influence_of_control_tokens_on_examples} illustrates this with an example.
The NbChars control token affects Levenshtein similarity: reducing the length decreases the Levenshtein similarity.
Finally, NbChars has a marginal impact on the WordRank ratio distribution, but clearly influences the dependency tree depth. This is natural considered that the depth of a dependency tree is very correlated with the length of the sentence.

The LevSim control token also has a clear cut impact on the Levenshtein similarity (row 2, col 2).
The first example in Table~\ref{influence_of_control_tokens_on_examples} highlights that LevSim increases the amount of paraphrasing in the simplifications. With an extreme target ratio of 0.25, the model outputs ungrammatical and meaningless predictions, thus indicating that the choice of a target ratio is important for generating proper simplifications.

WordRank and DepTreeDepth do not seem to control their respective attribute as well as NbChars and LevSim according to Figure~\ref{influence_of_control_tokens_with}.
However we witness more lexical simplifications when using the WordRank ratio than with other control tokens. In Table~\ref{influence_of_control_tokens_on_examples}'s first example, "designated as" is simplified by "called" or "known as" with the WordRank control token.
Equivalently, DepTreeDepth splits the source sentence in multiple shorter sentences in Table~\ref{influence_of_control_tokens_on_examples}'s first example.
WordRank and DepTreeDepth control tokens therefore have the desired effect.

\section{Conclusion}
This paper showed that explicitly conditioning Seq2Seq models on control tokens such as length, paraphrasing, lexical complexity or syntactic complexity increases their performance significantly for sentence simplification. We confirmed through an analysis that each control token has the desired effect on the generated simplifications.
In addition to being easy to extend to other attributes of text simplification, our method paves the way toward adapting the simplification to audiences with different needs. 

\section{Bibliographical References}
\label{main:ref}
\bibliographystyle{lrec}
\bibliography{bibliography}

\section{Language Resource References}
\label{lr:ref}
\bibliographystylelanguageresource{lrec}
\bibliographylanguageresource{bibliographylanguageresource}

\end{document}